# Saliency Detection in Educational Videos: Analyzing the Performance of Current Models, Identifying Limitations and Advancement Directions

Evelyn Navarrete
L3S Research Center, Leibniz
University Hannover
Hannover, Germany
navarrete@l3s.de

Ralph Ewerth
L3S Research Center, Leibniz
University Hannover
Hannover, Germany
TIB – Leibniz Information Centre for
Science and Technology
Hannover, Germany
ralph.ewerth@tib.eu

Anett Hoppe
L3S Research Center, Leibniz
University Hannover
Hannover, Germany
TIB – Leibniz Information Centre for
Science and Technology
Hannover, Germany
anett.hoppe@tib.eu

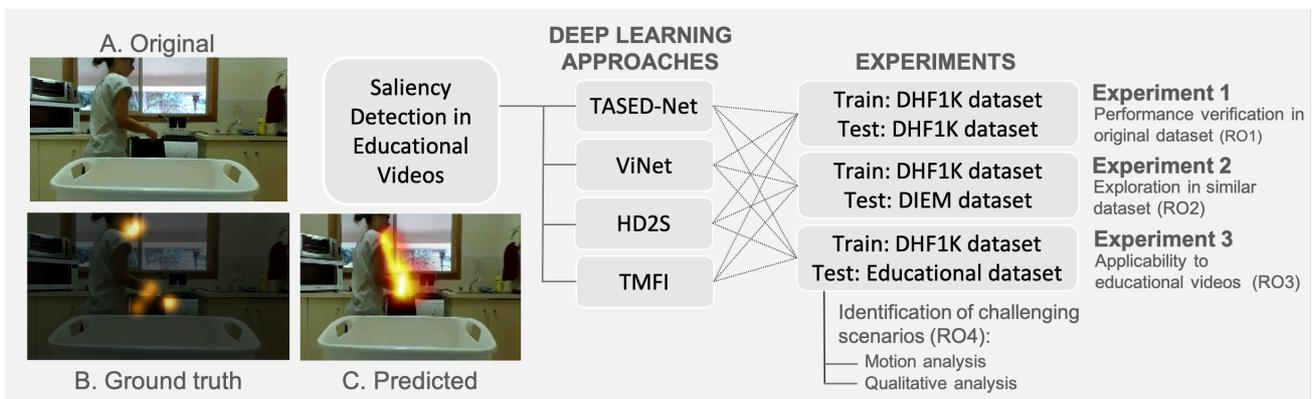

Figure 1: Experimental design overview and example of saliency prediction [25]. The original video frame is displayed in A, the ground truth in B (real eye fixations of viewers), the prediction in C, and saliency is highlighted in red-yellow. Original images in A, B, and C are from Wang et al. [25], licensed under CC BY 4.0, available at: Link

## ABSTRACT
Identifying the regions of a learning resource that a learner pays attention to is crucial for assessing the material's impact and improving its design and related support systems. Saliency detection in videos addresses the automatic recognition of attention-drawing regions in single frames. In educational settings, the recognition of pertinent regions in a video's visual stream can enhance content accessibility and information retrieval tasks such as video segmentation, navigation, and summarization. Such advancements can pave the way for the development of advanced AI-assisted technologies that support learning with greater efficacy. However, this task becomes particularly challenging for educational videos due to the combination of unique characteristics such as text, voice, illustrations, animations, and more. To the best of our knowledge, there is currently no study that evaluates saliency detection approaches in educational videos. In this paper, we address this gap by evaluating four state-of-the-art saliency detection approaches for educational videos. We reproduce the original studies and explore the replication capabilities for general-purpose (non-educational) datasets. Then, we investigate the generalization capabilities of the models and evaluate their performance on educational videos. We conduct a comprehensive analysis to identify common failure scenarios and possible areas of improvement. Our experimental results show that educational videos remain a challenging context for generic video saliency detection models.

## 1 INTRODUCTION
Video saliency detection (VSD) addresses the task of automatically locating regions that capture a viewer's attention, typically visual, in videos [15]. It is often operationalized as a prediction task where the ground truth is given by the eye fixations collected in user studies. An example is shown in Figure 1. The prediction of regions that capture viewers' attention poses significant challenges [5, 15]. However, thanks to the progress in deep learning architectures, there were significant advancements for VSD. Today, the leaderboards of popular benchmarks are dominated by deep learning approaches [1, 14], achieving remarkable results. Saliency detection in videos can enhance information retrieval (IR) tasks by assigning additional metadata to the index [15]. Particularly for video-based learning, it can support tasks such as video segmentation, navigation, and summarization which are the basis for video retrieval (e.g., [4, 18, 26]). Identifying salient regions, for example, improves video indexing by assigning region information to the corresponding video segments (e.g., [18]). Effective segmentation methods have been shown to optimize video data access and decrease retrieval times (e.g., [9]). However, the automatic saliency detection in educational videos presents significant challenges due to their unique mix of modalities such as text, voice, illustrations, and animations, which differ from non-educational videos used to train generic models.

Evaluating information retrieval and pattern recognition approaches not only in the same context as the original studies but also across both similar and dissimilar contexts remains a persistent challenge [11, 22]. However, this is not only desirable but necessary to ensure consistent results and conclusions [22]. This also applies to VSD, as the reliability of the methods also impacts the performance of downstream tasks and applications. To the best of our knowledge, there is no systematic study that evaluates state-of-the-art models in automated saliency recognition specifically for educational videos, and thus several research questions remain open: Is it possible to apply a model trained on data from other domains to educational videos? Can these methods, without further customization, serve as effective tools when data in the educational domain are unavailable or costly?

**Research objectives:** In this paper, we investigate the applicability of generic saliency detection methods to educational videos. For this purpose, we choose four state-of-the-art approaches based on their performance on the DHF1K benchmark dataset [1]. TASED-Net (temporally-aggregating spatial encoder-decoder network) [20], HD$^2$S (hierarchical decoding for dynamic Saliency) [3], ViNet (visual-only network that leverages skip connections, trilinear up-sampling) [12], and TMFI (transformer-based multi-scale feature integration network). Our research objectives are as follows: To ensure a reliable evaluation of the approaches, we first **(RO1)** verify the performance of the approaches as reported in the original studies and **(RO2)** explore their robustness and sensitivity via a replicability study with a different dataset from a similar context. Ultimately, we seek to **(RO3)** evaluate the applicability of the methods to an educational context, and **(RO4)** find typical scenarios where the methods fail to detect salient regions. An overview of the experimental design is shown in Figure 1.

**Contributions:** (1) To the best of our knowledge, this is the first study that evaluates the applicability of generic video saliency detection methods to educational video resources. (2) Our research advances the understanding of model applicability in the real-world setting of educational videos. (3) Our analysis of typical failure scenarios provides insights for enhancing VSD in educational applications, which could further contribute to a better search and retrieval in educational video databases.

The rest of this paper continues with a description of related work (Section 2), a description of the selected saliency detection approaches (Section 3), the experimental setup (Section 4), and the experimental results (Section 5). Finally, we draw conclusions, identify limitations and outline areas of future work (Section 6). For transparency, we share our source code (link to repository).

## 2 RELATED WORK

**Video saliency detection:** In this study, VSD is conceptualized as the task of predicting visual regions that draw the attention of viewers while watching videos, specifically aligned to the estimation of eye fixations (see Figure 1).

**Datasets:** There are three widely used datasets for VSD [5]: (1) DIEM [21], which contains 84 videos about sports and movie trailers, (2) Hollywood-2 [19], the largest dataset, includes 1,707 short videos of Hollywood movies, and (3) UCF-Sports [23], which includes 150 videos about sports. Moreover, two additional datasets have gained prominence [5]: (1) DHF1K [25], one of the biggest datasets with 1,000 videos on diverse categories related to human activities, motions, and scenes, and (2) LEDOV [13], a dataset with 538 videos about sports, daily actions, and social activities. DHF1K is notable for its diversity and complexity. The approaches selected for this work achieved the best performance in their original studies using this last dataset. This is the reason why it is further used as a base model to fine-tune on other domains. In this study, we have used the DHF1K [25] and DIEM [21] datasets. Table 1 presents more details.

**Video saliency detection for education:**

Identifying the regions of a learning resource that a learner pays attention to is a crucial task to understand how the material affects the learner. Studies on attention-drawing (i.e., salient) regions have deepened our understanding of how multimedia content influences factors such as learners' cognitive load, engagement, and learning outcome [2, 8, 17]. These insights are critical for assessing the quality of educational resources and to systematically improve their design. VSD has been applied to achieve different tasks such as automatically generating tables of contents for educational videos using saliency to detect the importance of words in on-screen slides [4, 26]. Other works [18] used saliency detection to identify key visual information and, thereby determine relevant video segments, which can be further organized into a table of contents to allow navigation and summarization.

## 3 METHODS

**Selection criteria:** We choose four state-of-the-art video saliency detection approaches based on: (1) Their performance as reported on the DHF1K benchmark [1], one of the largest and most used datasets for this task, and (2) the availability of the original source code – to avoid variations caused by reimplementation and missing parameters. These approaches are introduced below.

The **TASED-Net approach [20]** proposes a 3D fully convolutional encoder-decoder network architecture. It uses temporal aggregation inside the network to progressively reduce the dimension, which is achieved by introducing new pooling operations. The videos are encoded with a network pre-trained for action recognition. The decoder network uses transposed convolution layers and max-pooling layers for spatial up-scaling and convolutional layers for temporal aggregation. The hyper-parameters for the training process are set with batch sizes of 40, learning rate at 1e-3 for the encoder network and 0.1 for the decoder network, and Kullback-Leibler divergence as loss function.

The **HD$^2$S approach [3]** uses a 3D fully convolutional network encoder-decoder architecture. As TASED-Net, it uses a network pre-trained for action recognition to encode video frames. For the prediction, it introduces a hierarchical decoding that generates multiple saliency maps using features learned at different levels of abstraction. For the training process, a learning rate of 1e-3, $L_2$ regularization with weight decay, Adam as optimizer, 2500 training iterations, a batch size of 200, and Kullback-Leibler divergence as loss function are set.

The **ViNet approach [12]** presents a 3D fully convolutional encoder-decoder architecture. Again, it uses a network pre-trained for action recognition to encode the video frames. The decoder combines features from the multiple hierarchies to infer a saliency map. However, in contrast, this approach employs alternative techniques that leverage skip connections and trilinear up-sampling. For the training process, it is stated that the learning rate is 1e-4, Adam function is used as optimizer, and the Kullback-Leibler divergence metric as loss function.

**TMFI [28]** is currently the leading approach in the benchmark [1]. Different to the previous methods, it integrates a transformer technology with a semantic-guided encoder. This generates spatio-temporal features with high-level semantic information across multiple levels. The decoder refines the features using a multi-dimensional attention module. For the training process, the method sets an initial learning rate of 1e-5, batch size of 1, Adam as optimizer function, and Kullback-Leibler Divergence and Linear Correlation Coefficient as loss functions.

## 4 EXPERIMENTAL SETUP

### 4.1 Datasets

We have used the **DHF1K [25]** dataset to verify the performance of the approaches as reported in the original studies. To explore the robustness and sensitivity of the methods in a context similar to the original one, we have used the **DIEM [21]** dataset (see Section 2 for details).

To investigate the applicability of the methods to educational videos we assembled an **educational dataset** from Madsen et al. [17] and Zhang et al. [27]. The former was originally used to study how synchronized eye movements of learners watching educational videos can predict learning performance. From this study, data corresponding to two educational videos in animated style were suitable for our study. The dataset from Zhang et al. [27] originates from studying how mind-wandering is related to changes in the eye movements of learners while watching educational videos. Data corresponding to two videos in slide-based style, including the instructor embodiment, was available. Video frame samples of the videos in the educational dataset are displayed in Figure 4. Table 1 a summarizes the most important aspects of all these datasets.

**Datasets pre-processing:** The VSD task requires a particular representation of the input data. The ground truth is represented as **fixation maps** and fulfills the role of the labels for the training process. This map consists of a frame in gray-scale in the range from zero to 255 for each pixel where the brighter the pixel, the more salient it is considered. The fixation map is obtained by plotting eye-fixation coordinates, collected typically using eye-tracking software, into the frame and building a so-called density map computed with a 2D Gaussian function. This map reveals the distribution of the salient visual regions. Likewise, the resulting prediction yields a **saliency map** in gray-scale where the degree of saliency is directly proportional to the brightness of each pixel [15].

**Table 1: Summary of the datasets used in the experiments**

| Dataset name | # of videos | Video length | Video categories |
|---|---|---|---|
| DHF1K [25] | 1,000 | 19.42 secs. average per video | Daily and social activities, animals, scenery, sports |
| DIEM [21] | 84 | 20 to 200 secs. | Game trailers, sports, ads, music, news, movie trailers, documentaries |
| Educational [17, 27] | 4 | 150 to 1200 secs. | Physics, biology, education studies |

### 4.2 Experiments

**Experiment 1:** This experiment corroborates the performance of the TASED-Net [20], ViNet [12], HD$^2$S [3], and TMFI [28] models trained on the DHF1K dataset [25] as reported in their original publications (RO1). Note that the test set of the DHF1K dataset [25] is not openly available. Consequently, we employed either the validation set or an equivalent number of videos from the available dataset for method testing, consistent with the procedures of the original research. Since the original test set is not available, we investigate how well we the results can be reproduced. For all the models in this experiment, we ensured that the hyper-parameters and training procedures were the same (as summarized in Section 3). Moreover, to account for randomness, we have run this experiment in five iterations for each approach with no changes in the hyper-parameter settings and test set. This task helped us examine how sensitive the models are to random initializations.

**Experiment 2:** This experiment focuses on exploring the performance of the four approaches on the DIEM [21] dataset (RO2), which has a similar context to the original (DHF1K [25]). See a comparison of these datasets in Table 1. Through this examination, we explore a scenario where no data is available, both for training a customized model or fine-tuning an existing one. Consequently, we applied the four models as out-of-the-box solutions with no fine-tuning. As the dataset providers do not specify a test dataset, we use 17 randomly selected videos. Moreover, it is worth noting that the DIEM dataset was used for evaluation in the study proposing the ViNet [12] approach but not in the rest of the studies [3, 20, 28].

**Experiment 3:** In this experiment, we tested the four approaches with educational content. To achieve this, we used a dataset of educational videos (see details in Section 4.1) to evaluate the performance of each model. Similarly to the previous experiment, we did not fine-tune the models on this dataset.

### 4.3 Evaluation

**Evaluation Metrics:** The saliency detection approaches were evaluated on four metrics widely used to assess the performance on this task [7]: Pearson's correlation coefficient (CC), distribution similarity (SIM), normalized scanpath saliency (NSS), and Judd area under the curve variant (AUC-J). For an in-depth understanding of the calculation and limitations of these metrics, the reader is referred to Bylinskii et al. [7]. Our implementation is based on the source code provided by the MIT/Tuebingen Benchmark [14]

**Motion Estimation Analysis:** Low motion intensity can be a potential factor, decreasing the performance of video saliency detection methods (e.g., [3, 28]). Accordingly, we conducted a motion analysis in educational videos to determine whether low motion level entails a decreased saliency recognition performance. To this end, we investigated the relation between the level of motion in videos and the corresponding performance of saliency detection models applied to these videos. The motion level is estimated at video frame level by measuring the movement between consecutive frames. Specifically, we implemented the Farneback method [10] provided by the Open Source Computer Vision Library (OpenCV) [6], a library for computer vision tasks. This method compares two frames and generates a motion vector per pixel. The vector is composed of the direction and the intensity of the movement, where the last one can be represented as a magnitude. The magnitude can be interpreted as the velocity of the pixels' movement between frames, where a larger magnitude indicates a faster movement. For each pair of consecutive video frames, the magnitudes are computed per pixel and then averaged for the whole frame.

**Qualitative Analysis:** We conducted a manual examination of the predicted saliency of individual video frames in the educational dataset. This analysis aimed to identify the conditions or scenarios in which video saliency detection methods encounter challenges, thereby stressing possible areas of improvement.

## 5 RESULTS

This section reports the results of the three experiments: Performance verification in the original dataset, exploration in a similar dataset, and applicability to educational videos (see Section 4.2). Furthermore, we present the findings from the motion and qualitative analyses to identify the challenges educational videos pose to saliency detection methods.

### 5.1 Performance Verification (RO1)

The first experiment aims at reproducing and verifying the results of the VSD approaches, TASED-Net [20], ViNet [12], HD$^2$S [3], and TMFI [28] as reported in the original studies. The process is explained in Section 4.2. Table 2 reports a summary of our findings. It displays the results we obtained in each of the five iterations. We have further included the standard deviation ("STD") as well as the average ("AVG") across these multiple iterations. Concerning the standard deviations ("STD"), overall, they are exceedingly minimal for all the models. This leads us to consider that the models are relatively robust to the effect of random initializations. The average column ("AVG") represents our final results, which we compared against the original results ("Original"). The highest scores are highlighted in bold. For all four approaches, we can observe disparities between our results and the ones of the original studies. We attribute these differences primarily to the use of the other test set due to the unavailability of the original test set (see Section 4.2). Nonetheless, the results are very close to the original in all cases, we thus consider that the four approaches are reproducible.

### 5.2 Exploration in Similar Contexts (RO2)

Our second experiment replicated the experiments, using a slightly different context than the one presented in the original studies (see Section 4.2). For this task, we used the DIEM dataset (see Section 4.1 for details). Table 3 presents the results we obtained for each metric and respective approaches. Note that the study that proposed the ViNet [12] approach evaluates the model on the DIEM dataset [21] (without fine-tuning). Thus, only in this case, we have included the results of this specific evaluation ("Original*" column) to be compared with our results. The differences with respect to our results remain minor for the AUC-J and NSS metrics across all the approaches. Interestingly, the results of all the models vary notably for the CC and SIM metrics. In this case, replicability on the DIEM dataset yields notably higher results than those observed in the original studies. However, we detected a similar pattern in the ViNet [12] study. Upon comparing the ViNet [12] results evaluated in the DIEM dataset ("Original*") with their base results ("Original"), we observe how the CC and SIM metrics differ significantly in favor of the DIEM scenario, similar to our results. In any case, we think that these findings offer positive indications that these models can be applied to similar scenarios and, to a considerable degree, replicate their performance.

### 5.3 Applicability to Educational Videos (RO3)

The third experiment transferred the saliency detection models to a different usage context than the one(s) they were trained for. In this experiment, we used a dataset related to educational videos (see Sections 4.1 and 4.2 for details). The results are summarized in Table 4. For the AUC-J, CC, and NSS metrics, we can observe that the original studies mostly present higher scores compared to the application on educational videos. The performance of the models measured by the CC metric is slightly lower than in the original, moderately lower when measured by the AUC-J metric, and notably lower when measured by the NSS metric. On the other hand, our SIM scores are significantly higher in comparison to the original studies for two of the models (TASED-Net [20] and ViNet [12]). Nonetheless, given the generally lower performance observed throughout most metrics in the educational video context, it is evident that the saliency detection models do not exhibit the same level of discriminative capability as they do in their original context. Yet, it is important to note that despite this drop in performance, the models' performance in the educational context is not at its lowest point. This observation hints at the potential of these models to exhibit some degree of applicability to educational videos. In particular, the best performance across all metrics was achieved by ViNet [12].

Furthermore, we conducted an analysis of the results with regard to the production style. However, considering that the investigated models do not perform at a similar level within the educational domain, as compared to the original studies, our analysis concentrated only on the best predictions. We specifically analyzed the top 10, 20, 30, 40, and 50% of predictions to account for the instances where the models most accurately predicted saliency. The results are depicted in Figure 2 and show that in each of the five top-performing categories, the models were more effective in predicting saliency for animated styles than for slide-based styles. This distinction is particularly noticeable for metrics such as AUC-J and CC, which show a clear separation in performance between the two styles. One possible explanation for this difference could be variations in motion levels. We investigate this possibility further in Section 5.4.

Table 2: Performance verification of original studies on the DHF1K dataset [25] according to the metrics Pearson's correlation coefficient (CC), similarity (SIM), normalized scanpath saliency (NSS), and Judd area under the curve (AUC-J

| Model | Metric | 1st. | 2nd. | 3rd. | 4th. | 5th. | STD | AVG (our results) | Original results |
|---|---|---|---|---|---|---|---|---|---|
| TASED-Net | AUC-J | 0.897 | 0.896 | 0.898 | 0.895 | 0.896 | 0.0013 | **0.896** | 0.895 |
| | CC | 0.503 | 0.504 | 0.504 | 0.5009 | 0.504 | 0.0013 | **0.503** | 0.470 |
| | NSS | 2.849 | 2.848 | 2.869 | 2.849 | 2.869 | 0.0112 | **2.857** | 2.667 |
| | SIM | 0.391 | 0.393 | 0.392 | 0.392 | 0.393 | 0.0008 | **0.392** | 0.361 |
| ViNet | AUC-J | 0.899 | 0.900 | 0.899 | 0.903 | 0.899 | 0.0019 | 0.900 | **0.908** |
| | CC | 0.529 | 0.529 | 0.529 | 0.531 | 0.529 | 0.0008 | **0.529** | 0.510 |
| | NSS | 3.089 | 3.087 | 3.083 | 3.076 | 3.086 | 0.0045 | **3.084** | 2.870 |
| | SIM | 0.387 | 0.387 | 0.387 | 0.383 | 0.387 | 0.0018 | **0.386** | 0.381 |
| HD$^2$S | AUC-J | 0.894 | 0.892 | 0.896 | 0.897 | 0.897 | 0.0019 | 0.895 | **0.908** |
| | CC | 0.485 | 0.486 | 0.486 | 0.484 | 0.487 | 0.0012 | 0.486 | **0.503** |
| | NSS | 2.703 | 2.712 | 2.683 | 2.690 | 2.696 | 0.0112 | 2.697 | **2.810** |
| | SIM | 0.398 | 0.395 | 0.395 | 0.395 | 0.396 | 0.0015 | 0.396 | **0.406** |
| TMFI | AUC-J | 0.9079 | 0.8984 | 0.9082 | 0.9074 | 0.8987 | 0.0051 | 0.9041 | **0.9153** |
| | CC | 0.5700 | 0.5759 | 0.5695 | 0.5711 | 0.5660 | 0,0036 | **0.5705** | 0.5461 |
| | NSS | 3.3216 | 3.3707 | 3.3293 | 3.3138 | 3.3134 | 0.0238 | **3.3297** | 3.1460 |
| | SIM | 0.4311 | 0.4270 | 0.4180 | 0.4216 | 0.4286 | 0.0054 | **0.4253** | 0.4068 |

Table 3: Performance in a similar context on the DIEM dataset [21] according to the metrics Pearson's correlation coefficient (CC), similarity (SIM), normalized scanpath saliency (NSS), and Judd area under the curve (AUC-J)

| Metric | TASED-Net | | ViNet | | | HD$^2$S | | TMFI | |
|---|---|---|---|---|---|---|---|---|---|
| | DIEM | Original | DIEM | Original* | Original | DIEM | Original | DIEM | Original |
| AUC-J | 0.853 | **0.895** | 0.850 | 0.886 | **0.908** | 0.855 | **0.908** | 0.8642 | 0.9153 |
| CC | **0.634** | 0.470 | **0.638** | 0.571 | 0.510 | **0.608** | 0.503 | **0.6608** | 0.5461 |
| NSS | 2.573 | **2.667** | 2.635 | 2.280 | **2.870** | 2.429 | **2.810** | 2.7447 | **3.146** |
| SIM | **0.532** | 0.361 | **0.513** | 0.468 | 0.381 | **0.511** | 0.406 | **0.5319** | 0.4068 |

Table 4: Performance in the educational dataset (Ed.) [17, 27] according to the metrics Pearson's correlation coefficient (CC), similarity (SIM), normalized scanpath saliency (NSS), and Judd area under the curve (AUC-J)

| Metric | TASED-Net | | ViNet | | HD$^2$S | | TMFI | |
|---|---|---|---|---|---|---|---|---|
| | Ed. | Original | Ed. | Original | Ed. | Original | Ed. | Original |
| AUC-J | 0.827 | **0.895** | 0.838 | **0.908** | 0.810 | **0.908** | 0.8012 | **0.9153** |
| CC | 0.456 | **0.470** | **0.521** | 0.510 | 0.481 | **0.503** | 0.4413 | **0.5461** |
| NSS | 1.629 | **2.667** | 1.734 | **2.870** | 1,542 | **2.81** | 1.6654 | **3.146** |
| SIM | **0.482** | 0.361 | **0.510** | 0.381 | **0.434** | 0.406 | **0.4575** | 0.4068 |

### 5.4 Identification of Challenging Scenarios (RO4)

In this section, we identify scenarios that challenge saliency detection for educational videos. We have conducted both a motion and a qualitative analysis.

**Motion Estimation Analysis:** Prior studies noted that their models struggle to accurately detect saliency in videos characterized by minimal motion [3, 28]. To investigate this, a motion analysis was conducted. We first computed the motion level of the educational videos as described in Section 4.3. Table 5 presents basic statistics concerning the motion level of the videos in the educational dataset [17, 27] in contrast to the videos in the test dataset of the original studies (DHF1K [25]) [3, 12, 20, 28].

Table 5: Motion level (average, standard deviation, maximum, minimum) of original [25] and educational datasets [17, 27]

| Dataset | Avg. | Std. | Max. | Min. |
|---|---|---|---|---|
| **Educational** | 0.334 | 0.366 | 0.830 | 0.008 |
| - *Slide-based* | 0.063 | 0.079 | 0.119 | 0.008 |
| - *Animated* | 0.605 | 0.319 | 0.830 | 0.379 |
| **Original** | 1.777 | 1.800 | 9.127 | 0.027 |

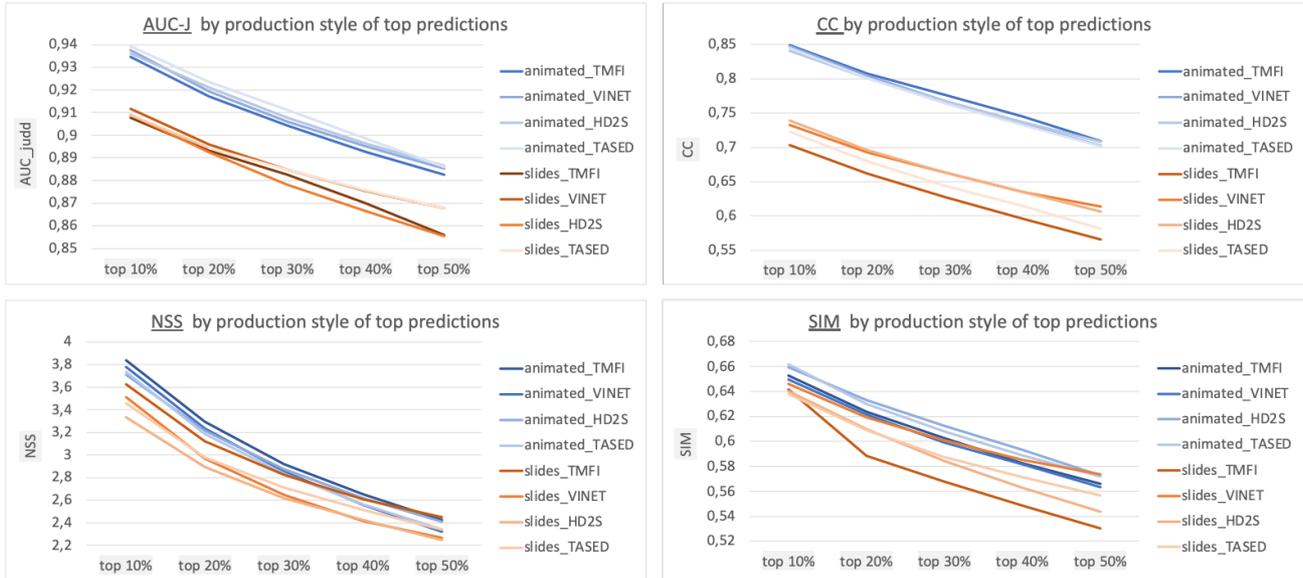

Figure 2: Performance of the top predictions according to video production styles as measured by Pearson's correlation coefficient (CC), similarity (SIM), normalized scanpath saliency (NSS), and Judd area under the curve (AUC-J) metrics

The motion level of the videos in our educational dataset in terms of average, standard deviation, maximum, and minimum is substantially lower than in the original dataset. This decrease is particularly notable in slide-based styles versus animated styles. Consequently, we anticipated a tendency of decreasing performance of the detection models as the motion in our educational videos decreases. To examine this conjecture, we examined the relation between motion level and model performance across the educational videos. Similar to the previous section, our focus remained on the top 10, 20, 30, 40, and 50% of the best predictions in order to study only the instances where the models were most successful at predicting saliency. The results are displayed in Figure 3.

Our findings show a clear trend where lower motion levels correspond to less accurate predictions when performance is evaluated using AUC-J and NSS metrics. A similar but more moderate trend is observed with the CC metric from the top 20 to 50% of the best predictions. The only discrepancy is presented with the SIM metric, which shows a slight increase in motion as accuracy decreases. Given that most metrics present a downward trend in performance as motion decreases, we conclude that motion levels can impact the effectiveness of saliency detection models in the educational videos dataset. Notably, videos in the slide-based style, which have lower motion levels compared to animated styles, would be more challenging for the models.

**Qualitative Analysis:**

To identify further cases under which video saliency detection methods face challenges, a manual examination was conducted. It focused on saliency predictions at the individual frame level (saliency maps) for the entire educational dataset. Given that the ViNet method [12] demonstrated the best performance in our educational dataset (see Section 5.3), we subsequently focused on the predictions of this approach. We manually analyzed 10% of the frames, choosing those presenting with the lowest performance scores over all metrics. Frames that repeatedly emerged were identified as posing particular challenges to video saliency detection.

Upon the examination, we identified the following scenarios (see also the corresponding examples in Figure 4). In **slide-based production styles**: (a) and (b) visual representation vs. human face: The model tends to predict higher saliency for human faces in comparison to other forms of information (i.e., figures and tables) even though these might have a higher importance in certain contexts and moments; (c) Table vs. text: The model predicts that text regions (e.g., paragraph) draw more attention than the content of a table that is more important in that context; (d) Table and text vs. face: The model predicts the human face as a more salient region than the table and text; (e) Text vs. image: The model predicts that an image that is not contextually relevant at the time is more salient than a relevant text region, and (f) Text regions: The model struggles to accurately predict which text region should draw attention among all the text regions. In **animated production styles** we observed the following challenges: (g) Complex backgrounds: The model struggles when presented with high-detail backgrounds, or very similar-looking imagery. (h) Shot and scene transitions: The model struggles during scene transitions, i.e., cuts or fades to new shots or scenes. On the other hand, we identified the following scenarios where salient regions were predicted accurately: (A) central regions: When a salient region is located at the center of the video frame, and (B) single line text: When a salient region corresponds to a text displayed as a single line.

## 6 DISCUSSION AND CONCLUSIONS

In this paper, we have investigated the usefulness of generic video saliency detection methods in the context of educational videos.

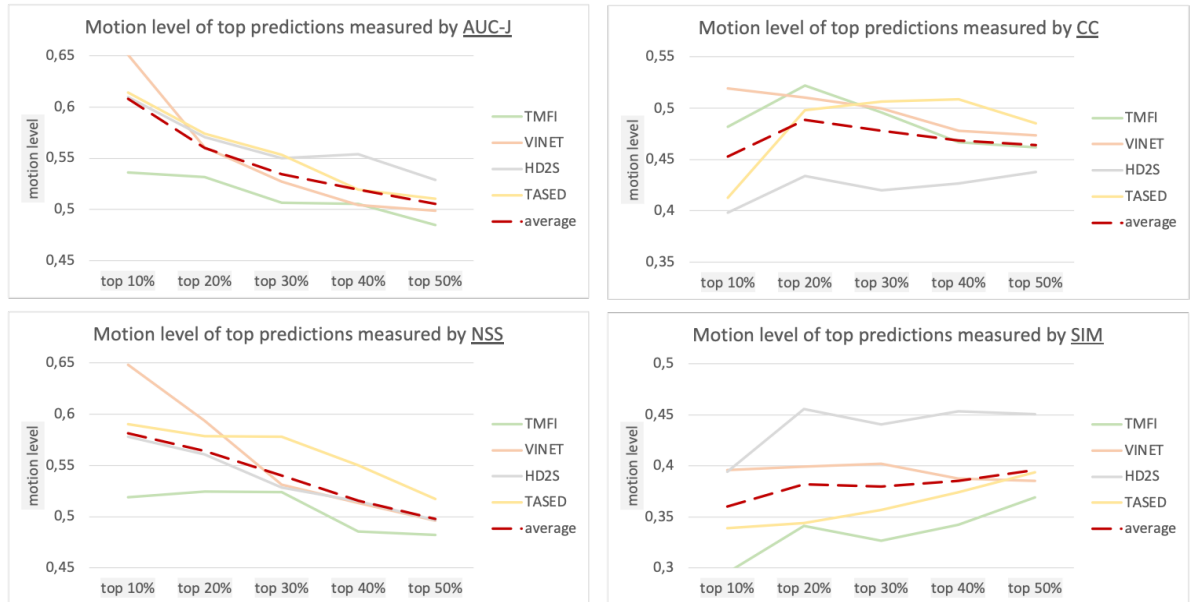

Figure 3: Motion level of the top predictions as measured by Pearson's correlation coefficient (CC), similarity (SIM), normalized scanpath saliency (NSS), and Judd area under the curve (AUC-J) metrics

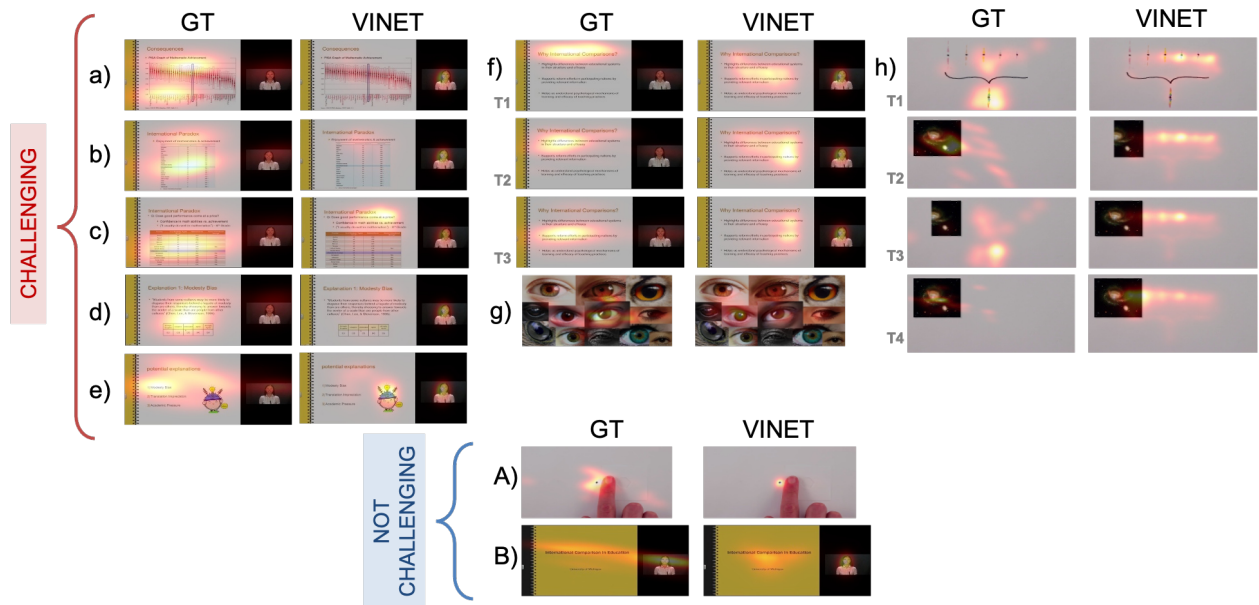

Figure 4: Scenarios where the ViNet [12] method failed to detect correctly salient regions. *GT* is the ground truth and $T_n$ denotes various sequential points in time. Original images in a-f, and A are from Zhang et al. [27], licensed under CC BY 4.0, available at: Link. Original images in g, h, and A courtesy of Neptune Studios LLC [16], used with permission.

Specifically, we have examined four state-of-the-art deep learning-based approaches: TASED-Net [20], HD$^2$S [3], ViNet [12], and TMFI [28]. We have addressed the following tasks: (1) We corroborated the performance of the approaches as stated in the original articles, (2) we examined their robustness and sensitivity by evaluating their performance in a different dataset with a similar context, (3) we explored the applicability of the models on educational videos, and (4) we identified typical scenarios where the approaches fail.

We were able to verify the results for the four approaches and also replicate their performance in a similar dataset to a considerable degree. However, although we could show that these models can be adapted to educational videos to some extent, their applicability remains a challenge.

**Challenges in domain transfer:** As the literature suggests that low-motion scenarios could challenge saliency detection methods [3, 28], we analyzed model performance based on the motion level of educational videos. Indeed, our motion analysis indicates that decreased motion is related to a decline in model performance. Moreover, a manual qualitative analysis of challenging video frames revealed that in **slide-based videos**, the models often struggled to determine *which* of the elements were contextually relevant. In particular, the model showed a tendency to attribute more importance to human faces than to other visual information such as text, tables, and figures. Further issues arose from correctly identifying the most relevant one from multiple sources of information, i.e., when confronted with several text passages, tables, or figures. Further issues arose from correctly identifying the most relevant element among multiple similar options, for example, when confronted with several text passages, tables, or figures. In **animated videos**, the methods struggled with (a) complex or high-detail backgrounds, and (b) scene transitions. In summary, we find that generic saliency detection methods struggle to perform on educational videos. It appears that characteristics of educational media differ too much from the original training material to enable a direct transfer. Indeed, previous research pointed out that the approaches underperform in cases featuring graphic designs, visualizations, and cartoon-like animations [5], which are often found in educational videos. This issue has been confirmed in our study. Moreover, past research detected a trend to emphasize text and human figures, even when there are other more semantically relevant aspects [5]. Also, it was identified that elements may compete for visual attention, potentially leading to poor model performance [5], a behavior also verified by this study. We extend these insights with regard to attention-competing elements in educational videos: The models show issues identifying the most relevant elements among instructor, tables, text, and figures.

**Limitations and future research directions:** Due to limited labeled data for educational videos, our study did not explore fine-tuning techniques. These strategies could improve model adaptation to educational contexts. Moreover, future research could consider other production styles apart from the slide-based and animated styles explored in this study. Additionally, we anticipate potential challenges for saliency detection approaches that may require further investigation, based on insights from prior research on generic models [5]: Models struggle in accurately detecting salient parts of a person and face, thus, videos featuring mainly the figure of the instructor might be challenging. Models tend to under/overestimate fixation location due to gaze direction. This might be a difficulty since instructors fix their gaze on specific parts of, for example, a slide as a cue for the learner to identify relevant information. Models also tend to fail in detecting saliency in profile faces, which can be challenging for classroom videos where the instructor switches between a blackboard and the audience. First-person videos are usually not included in generic datasets, yet they are common in educational contexts where the instructors record themselves solving exercises on a sheet of paper.

We highlighted examples of applications where automatic saliency detection could support video-based learning [4, 18, 26]. However, we see potential for new applications. For instance, Sharma et al. [24] demonstrated that guiding learners by displaying in the video the path that their gaze should follow can improve learning outcomes and sustain attention longer. Advancements in video saliency detection could automate such aids, fostering better learning environments.

## ACKNOWLEDGMENTS
This work has been partly supported by the Ministry of Science and Education of Lower Saxony, Germany, through the PhD Training Program "LernMINT: Data-assisted classroom teaching in the STEM subjects".